\documentclass{article}

\usepackage{microtype}
\usepackage{graphicx}
\usepackage{subcaption}
\usepackage{booktabs} 
\usepackage{algorithmic}
\usepackage{comment}
\usepackage{hyperref}
\usepackage{xspace}
\usepackage{paralist}

\newcommand{\sys}{CPR\xspace}

\newcommand{\pls}{PLS\xspace}

\usepackage{amsmath}
\usepackage{amssymb}

\usepackage[accepted]{mlsys2020}

\mlsystitlerunning{CPR: Understanding and Improving Failure Tolerant Training for Deep Learning Recommendation with Partial Recovery}

\begin{document}

\twocolumn[
\mlsystitle{CPR: Understanding and Improving Failure Tolerant Training for Deep Learning Recommendation with Partial Recovery}




\begin{mlsysauthorlist}
\mlsysauthor{Kiwan Maeng}{fb,cmu}
\mlsysauthor{Shivam Bharuka}{fb}
\mlsysauthor{Isabel Gao}{fb}
\mlsysauthor{Mark C. Jeffrey}{fb}
\mlsysauthor{Vikram Saraph}{fb}
\mlsysauthor{Bor-Yiing Su}{fb}
\mlsysauthor{Caroline Trippel}{fb}
\mlsysauthor{Jiyan Yang}{fb}
\mlsysauthor{Mike Rabbat}{fb}
\mlsysauthor{Brandon Lucia}{cmu}
\mlsysauthor{Carole-Jean Wu}{fb}
\end{mlsysauthorlist}

\mlsysaffiliation{fb}{Facebook AI}
\mlsysaffiliation{cmu}{Carnegie Mellon University}

\mlsyscorrespondingauthor{Kiwan Maeng}{kmaeng@andrew.cmu.edu}
\mlsyscorrespondingauthor{Carole-Jean Wu}{carolejeanwu@fb.com}


\vskip 0.3in

\begin{abstract}
The paper proposes and optimizes a \emph{partial recovery} 
training system, \sys, for recommendation models.
\sys relaxes the 
consistency requirement by enabling
non-failed nodes to proceed without
loading checkpoints when a node fails
during training, improving failure-related overheads.
The paper is the first to the extent of our knowledge to perform a data-driven, in-depth analysis of applying partial recovery to recommendation models and identified a trade-off between accuracy and performance.
Motivated by the analysis, we present \sys, a partial recovery 
training system that can reduce the training time and maintain the desired level of model 
accuracy by (1) estimating the benefit of partial recovery, (2) selecting an appropriate checkpoint saving interval, and (3) prioritizing to save updates of more frequently accessed parameters.
Two variants of \sys, \sys-MFU and \sys-SSU, reduce the checkpoint-related overhead from 8.2--8.5\% to 0.53--0.68\% compared to full recovery, on a configuration emulating the failure pattern and overhead of a production-scale cluster.
While reducing overhead significantly, \sys achieves model quality on par with the more expensive full recovery scheme, training the state-of-the-art recommendation model using Criteo’s Ads CTR dataset.
Our preliminary results also suggest that \sys can speed up training on a real production-scale cluster, without notably degrading the accuracy.
%
\end{abstract}
]
\printAffiliationsAndNotice{}

\section{Introduction}

Personalization and recommendation algorithms form the core of many internet services.
The goal of a personalized recommendation algorithm is to recommend content to users by balancing the probability of interaction with meaningful engagement. 
Personalized recommendation algorithms power products 
that suggest music on Spotify~\cite{spotify}, videos on YouTube and Netflix~\cite{youtube, netflix}, mobile applications on Google PlayStore~\cite{widedeep}, stories on Instagram~\cite{instagram},
commercial products~\cite{amazon, mckinsey}, or advertisements~\cite{baidu}.
The impact of recommendation algorithms on user experience is tremendous. 
Recent studies show that a significant amount of content---60\% of the videos on YouTube and 75\% of the movies on Netflix that were viewed---come from suggestions made by recommendation algorithms~\cite{mckinsey, underwood2019use, xie2018personalized}. 
The quality of the recommended content directly translates into service subscription~\cite{netflix} and revenue~\cite{baidu, mckinsey}.
Companies devote significant resources to recommendation models.
Across clusters servicing machine learning workloads, about 50\% of training and 80\% of inference cycles are dedicated to recommendation models at Facebook~\cite{arch_impl}.

Over the past decades, a plethora of research has been devoted to the development of recommendation algorithms, from classical techniques~\cite{contentbased, cf, mf} to machine learning~\cite{ncf, gbdt, pairwise} and deep learning~\cite{widedeep, dlrm, din, dien, deepfm, automated_interaction, featurehashing}.
Domain-specific systems tailored to deep learning-based recommendations have also been designed to enable high-performance and energy-efficient execution~\cite{baidu, aibox, intel, merlin, hugectr, tpu, a100, recnmp, centaur, tensordimm}. 

State-of-the-art recommendation models consist of two major components: Multi-Layer Perceptron layers
(MLP) and embedding layers (Emb), jointly trained to reach a target model quality~\cite{dlrm, widedeep}. 
MLP layers are replicated across multiple nodes (trainers) and run in parallel, while Emb layers are sharded across multiple embedding parameter server nodes (Emb PS) due to their large size~\cite{baidu}.
As the size and the complexity of recommendation models grow, the scale of MLP trainers and Emb PS nodes increases quickly, also increasing the expected failure rate of a multi-day/week recommendation model training.
By analyzing a large collection of industry-scale recommendation training jobs in production datacenter fleets with failures, our study finds that the mean-time-between-failures (MTBF) was \emph{14--30 hours} on average, similar to the statistics from other production-scale systems~\cite{failure_cmu, failure_fta, failure_google, failure_watson}.

A common approach to handle failures for distributed recommendation model training is with \emph{checkpointing}. 
A checkpointing system periodically saves the system state, or a checkpoint, to persistent storage.
At a failure, all nodes participating in the training load the last checkpoint, setting the model state back to a consistent, earlier version of the model. We refer to this baseline as \emph{full recovery}.
We observed that the overheads coming from checkpoints are non-negligible.
Our analysis of the production cluster shows that checkpoint-related overheads in full recovery can consume \emph{an average of 12\% of the total training time}. For the worst 5\%, \emph{training time slowdown can be up to 43\%}.
This 12\% overhead can add up to a significant
computational cost at scale.
By analyzing 17,000 training jobs from
a 30-day window, we observed that \emph{1,156 machine-years} worth
of computation was spent solely for failure handling.  

%
%
%
%

We propose to leverage \emph{partial checkpoint recovery} to improve the efficiency and reliability of recommendation model training.
Unlike full recovery, partial recovery restores a checkpoint only for the failed node, allowing all other nodes (trainer and Emb PS) to proceed with training without reverting their progress. 
Prior work showed that the iterative-convergent nature of ML training can successfully train the model
around the inconsistencies partial recovery introduces \emph{to a certain degree}~\cite{scar}.

However, we demonstrate in this paper that a naive application of
partial recovery can still harm the final model accuracy.
We identified that varying the checkpoint saving interval trades off the final model accuracy and the training time overhead in partial recovery, which is a unique, unexplored trade-off space.
%
%
To our knowledge, we are the first to conduct a thorough characterization study to understand this trade-off in the context of production-scale recommendation model training.
From the characterization study, we formulate a metric, 
\emph{portion of lost samples (\pls)}, to navigate through the trade-off space.

Using \pls,
we introduce \emph{Checkpointing with Partial recovery for Recommendation systems (\sys)}, the first partial recovery system that is customized for a large-scale production recommendation system training.
With the user-specified target \pls mapping to a certain accuracy target, \sys assesses the benefit of using partial recovery and selects a checkpoint saving interval to meet the target \pls with minimal overheads.
\sys implements two additional optimizations,
\emph{Most-Frequently Used} checkpointing (\sys-MFU) and \emph{Sub-Sampled Used} checkpointing (\sys-SSU), to further improve the accuracy when using partial recovery.
\sys-MFU and \sys-SSU leverage an important observation---
\emph{frequently accessed rows in the embedding table experience larger updates that have heavier effects when lost}.
\sys-MFU and \sys-SSU save updates of frequently accessed rows with higher priority under a constrained system bandwidth, thereby minimizing the degree of the model inconsistency introduced by failures.
%

We design, implement, and evaluate \sys
on (1) an emulation framework using the open-source MLPerf recommendation benchmark~\cite{mlperf} and (2) a production-scale cluster, training production recommendation models.
Our evaluation results show that \sys effectively reduces overheads while controlling the accuracy degradation using \pls.
On a framework emulating the production-scale cluster, we show \sys can reduce the checkpoint-related overheads by 93.7\% compared to full recovery, while only degrading accuracy by at most 0.017\%.
Also, our early results on a real-world production cluster demonstrate promising overhead reduction from 12.5\% to a marginal 1\%, without any accuracy degradation.
Our main contributions are:
\begin{compactitem}
    \item We perform the first in-depth, systematic analysis
    on the impact of partial recovery on recommendation model training.
    \item We introduce \pls,
    a novel metric that can be used to accurately predict the effect of partial recovery on model accuracy (Section~\ref{sec:sys}).
    \item We propose \sys, a practical partial recovery system
    for immediate adoption in real-world training systems (Section~\ref{sec:sys}).
\end{compactitem}

\section{Background and Motivation}
\label{sec:bg}

\subsection{Deep Learning Recommendation Systems}
\label{sec:bg_recsys}

\begin{figure}
    \centering
    \includegraphics[width=0.48 \textwidth]{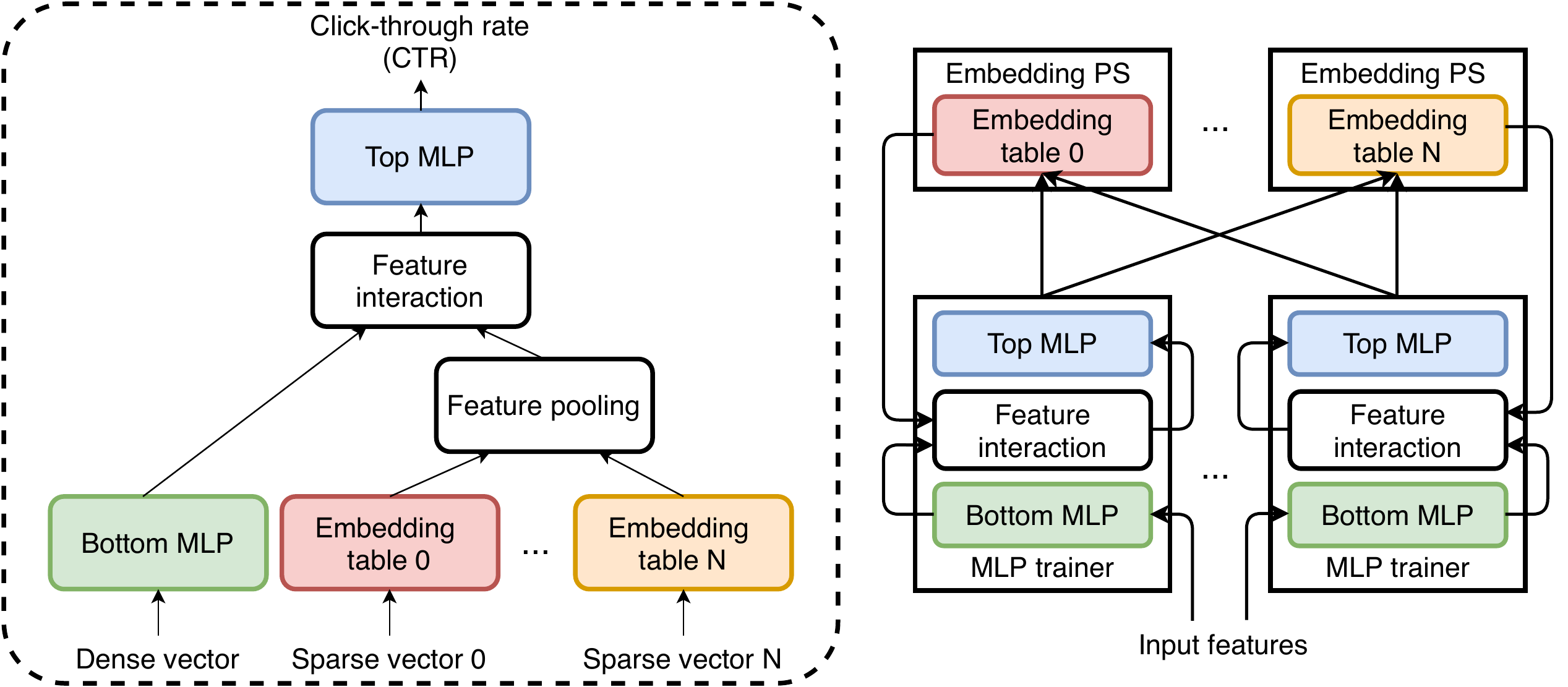}
    \vspace{-0.5cm}
    \caption{High-level overview of the recommendation
    system model architecture (left) and
    a typical training setup (right).}
    \vspace{-0.5cm}
    \label{fig:recsys_overview}
\end{figure}

Recommendation systems improve user experience by maximizing meaningful interactions. At a high level, a recommendation system takes user features, e.g., age or recently viewed videos, and content features, e.g., genre of movies or books, to predict whether the user
is likely to interact with certain contents.
Figure~\ref{fig:recsys_overview} depicts the generalized architecture for deep learning recommendation systems.  

There are two important feature types that are modeled in a recommendation system: \emph{dense} and \emph{sparse} features.
\emph{Dense features} represent
continuous inputs, such as user ages, that are used directly as inputs to the \emph{bottom MLP layer}.
\emph{Sparse features} represent categorical
inputs, such as movies or books a user has liked.
The sparse features are often encoded as multi-hot vectors, with only the indices mapping to a certain category set.
Because of the large feature domain, the multi-hot vectors are sparse---a few liked items among millions of items.  

Before being used, the sparse feature representation must go through \emph{embedding tables} and be translated to a dense vector representation. 
The \emph{embedding tables} can be viewed as lookup tables, where each row holds a dense \emph{embedding vector}.
Embedding vectors encode the semantics of each feature, and the distance between embedding vectors represents semantic relatedness.
%
The hot indices in the sparse feature representation are used as lookup indices to retrieve a set of embedding vectors, which are then combined in a
\emph{feature pooling layer} by operations such as
summation or multiplication.
The combined embedding vectors and the output of the bottom MLP layer are aggregated in the \emph{feature interaction layer}, where their similarity is calculated, e.g., with dot-product.
Finally, the result is fed into the 
{\it top MLP} layer, which predicts the likelihood of user engagement for the input user-item pair, i.e., \emph{click-through-rate or CTR}.


MLP layers are \emph{compute-intensive} and can be on the 
order of MBs in size. To exploit \emph{data-level parallelism}, an MLP layer is often replicated across multiple trainers and trained in parallel with different sets of data samples. 
Trainers synchronize their replicated parameters periodically, either through an MLP parameter server (not shown in the figure)~\cite{ps, easgd} or by point-to-point communication~\cite{sgp, 1bit}.

Embedding tables, on the other hand, are \emph{memory intensive}.
The embedding tables of production-scale recommendation models are often in the order of several hundreds of GBs to TB~\cite{baidu} in size and do not fit in a single-node training system. 
Thus, embedding table training exploits \emph{model-parallelism}, where tables are partitioned across multiple embedding parameter server (Emb PS) nodes and are jointly trained with all training data samples. 

\subsection{Checkpointing for Distributed Model Training}
\label{sec:bg_chkpt}

A common practice to handle failures in a distributed system is to periodically save a checkpoint, i.e., store a snapshot of the system state to persistent storage.
Checkpoints hold system states necessary to restore the progress. For ML training, checkpoints usually include the model parameters, iteration/epoch counts, and the state of the optimizer.
When any of the nodes fails, loading checkpoints for all the nodes (i.e., full recovery) reverts the system to the same state as when the checkpoint was saved.

There are four major overheads when using full recovery: (1) checkpoint saving overhead
($O_{save}$),
(2) checkpoint loading overhead
($O_{load}$), 
(3) lost computation
($O_{lost}$), 
and (4) rescheduling overhead ($O_{res}$).
Checkpoint saving/loading overhead refers to the time spent on saving/loading the checkpoint.
Lost computation is the amount of computation
executed between the last checkpoint and a failure.
Because the intermediate results were not saved, the same computation has to be re-executed.
Rescheduling overhead is the time spent for the cluster scheduler to find alternative, available nodes to take over the role of the failed nodes~\cite{condor, slurm}.

With an average node failure period $T_{fail}$ and the checkpoint saving interval $T_{save}$, a system's total overhead $O_{total}$ can be represented roughly as the following formula:
\begin{equation}
    O_{total} \approx O_{save}\frac{T_{total}}{T_{save}} + (O_{load} + \frac{T_{save}}{2} + O_{res})\frac{T_{total}}{T_{fail}}
    \label{eq:overhead}
\end{equation}
The first term ($O_{save}\frac{T_{total}}{T_{save}}$) represents the checkpoint saving overhead, calculated by multiplying the overhead of saving a checkpoint $O_{save}$ with the number of saving throughout training $\frac{T_{total}}{T_{save}}$.
Similarly, the second, third, and fourth terms represent the overhead of checkpoint loading, lost computation, and rescheduling, multiplied by the number of failures ($\frac{T_{total}}{T_{fail}}$).
Note that $O_{lost}=\frac{T_{save}}{2}$, assuming uniform failure probability.
The formula assumes each overhead is small compared to $T_{total}$. Otherwise, e.g., the number of checkpoint saving would have to be calculated as $\frac{T_{total} + O_{total}}{T_{save}}$ instead of  $\frac{T_{total}}{T_{save}}$.
With knowing the system parameter $O_{save}$, $O_{load}$, $O_{res}$, and
$T_{fail}$, the optimal checkpoint saving
interval $T_{save}$ that minimizes $O_{total}$ can be calculated: $T_{save,full} = \sqrt{2 O_{save}T_{fail} }$.

In a recommendation model training, Emb PS nodes account for most of the checkpoint-related overhead.
Unlike MLP layers that are small and replicated across trainers, embedding tables are large and partitioned across multiple nodes.
Thus, saving embedding tables is slow and requires coordination.
Conventional checkpointing strategies~\cite{checkpointing},
therefore, are inefficient for handling Emb PS failures---the optimization focus of this work. 

\subsection{Partial Recovery}
\label{sec:bg_pr}

As an alternative to full recovery, the concept of \emph{partial recovery} was proposed in recent work~\cite{scar}.
A distributed system with partial recovery only loads the checkpoint for the failed node, while keeping the progress of the remaining nodes.
%
%
Unless the iteration/epoch count is lost, partial recovery does not revert the progress, eliminating the need to re-execute computations.
The overhead of partial recovery does not contain the lost computation:
\begin{equation}
    O_{total\_par} \approx O_{save}\frac{T_{total}}{T_{save}} + (O_{load} +  O_{res})\frac{T_{total}}{T_{fail}}
    \label{eq:overhead_par}
\end{equation}
The performance benefit, however, comes at the expense of potential model quality degradation, because partial recovery introduces state inconsistencies across nodes.
Prior work~\cite{scar} proposes to compensate for the accuracy loss with training for additional epochs, although there is no guarantee on eventual model quality recovery.

\begin{figure}
    \centering
    \includegraphics[width=0.4\textwidth]{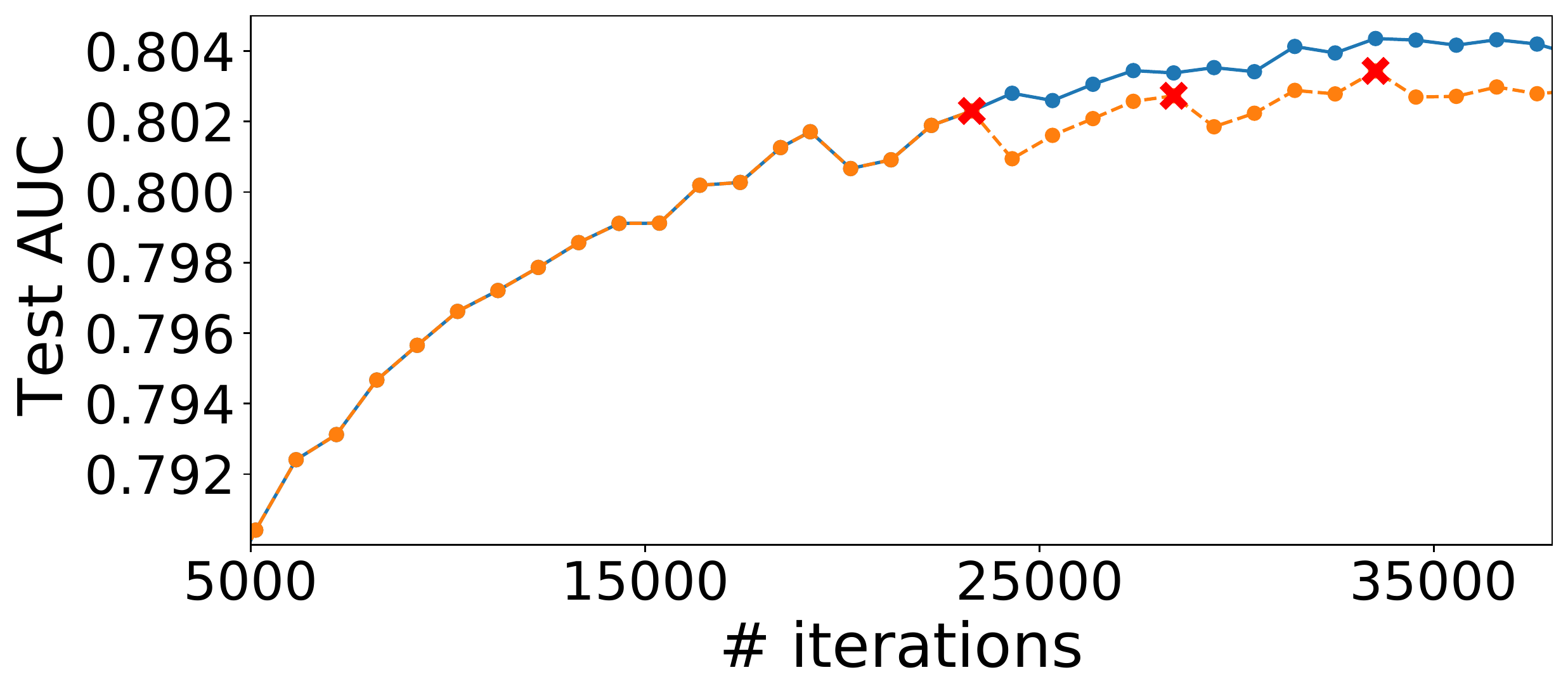}
    \vspace{-0.25cm}
    \caption{The accuracy of partial recovery system with failures (red crosses) never
    reached that of a non-failure
    case.
    }
    \vspace{-0.5cm}
    \label{fig:pr_bad}
\end{figure}

In fact, training for additional epochs does not always recover the model quality for recommendation model training, because recommendation models are prone to model overfitting when trained with more than one epoch~\cite{din}.
We show a motivational training scenario showing that partial recovery for recommendation model training can lead to irrecoverable accuracy degradation.
In Figure~\ref{fig:pr_bad}, failures (red cross) during training were handled by partial recovery (orange, dashed).
With partial recovery, the best accuracy is far lower than that without failures (blue, solid). Additional epochs do not close the accuracy gap, because recommendation models overfitted after the first epoch.
The experimental setup for Figure~\ref{fig:pr_bad} is discussed in Section~\ref{sec:eval}.

\paragraph{Unexplored design trade-off.}
The accuracy degradation of partial recovery can be potentially mitigated by saving checkpoints more frequently.
The relationship reveals a new trade-off space for partial recovery to explore--in partial recovery, changing the checkpoint saving interval trades off the training time overhead and model accuracy.
The role of the checkpoint saving interval for partial recovery is very different from that of full recovery, where the optimal value is simply $T_{save,full} = \sqrt{2 O_{save}T_{fail} }$.
Understanding the trade-off space is essential for the practical adoption of partial recovery on real-world applications.

%
\section{Understanding Failures for Production-Scale Training}
\label{sec:analysis}

Nodes in a large-scale training can fail for various reasons:
hardware failures~\cite{four_years, ares, machine_failures, ssd_failure, dean}, system failures (e.g., out-of-memory), user errors (e.g., bug in the code), and maintenance failures (e.g., kernel update)~\cite{googlecluster}.
While the failure probability of each node may be low, as the number of participating nodes increases, the likelihood of failure becomes higher.
%


\subsection{Failures for Distributed Recommendation Training}
\label{sec:bg_failure}

\begin{figure}
     \centering
     \begin{subfigure}[b]{0.21\textwidth}
         \centering
         \includegraphics[width=\textwidth]{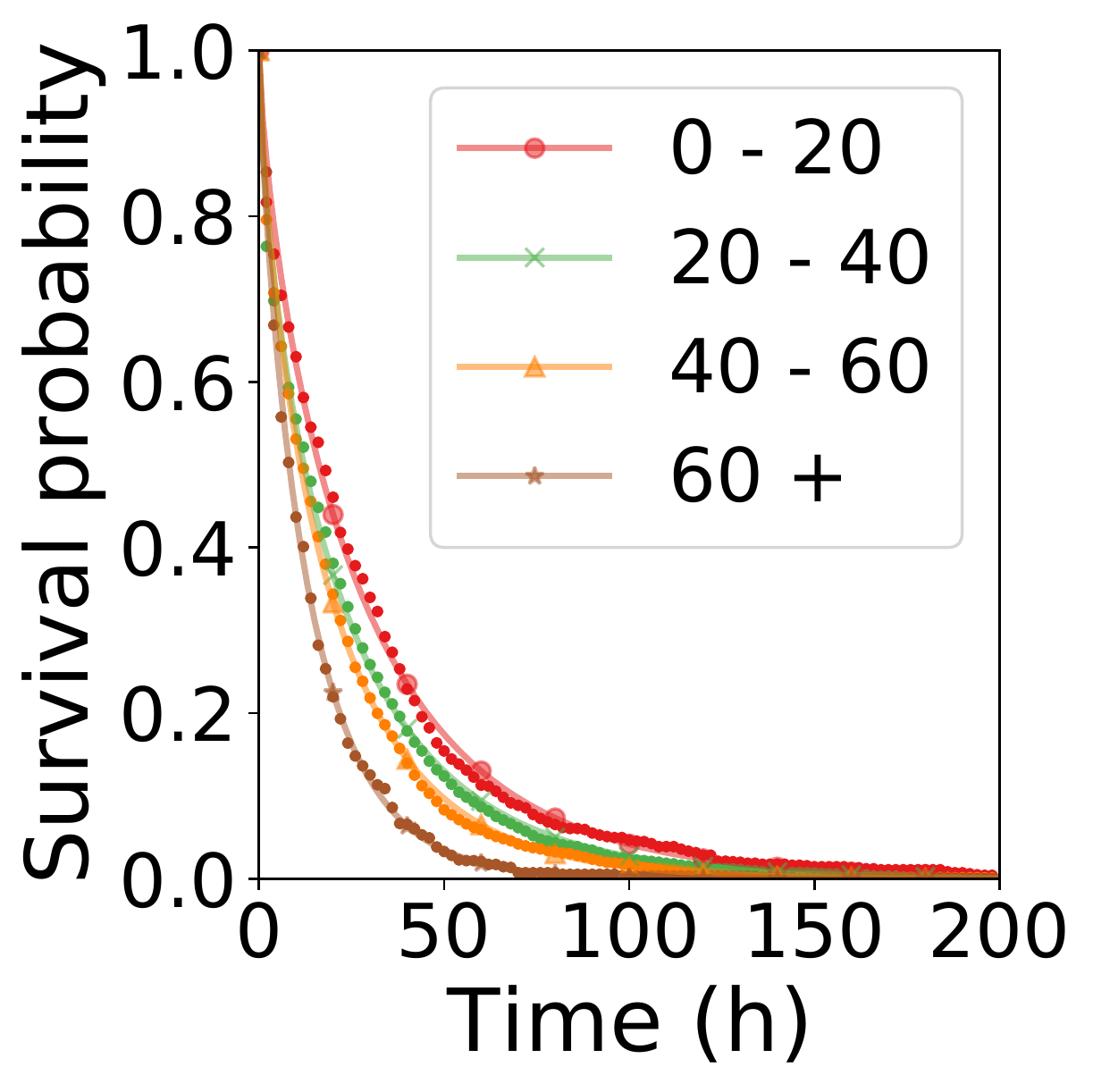}
         \caption{Survival distribution.}
         \label{fig:surv}
     \end{subfigure}
     \begin{subfigure}[b]{0.21\textwidth}
         \centering
         \includegraphics[width=\textwidth]{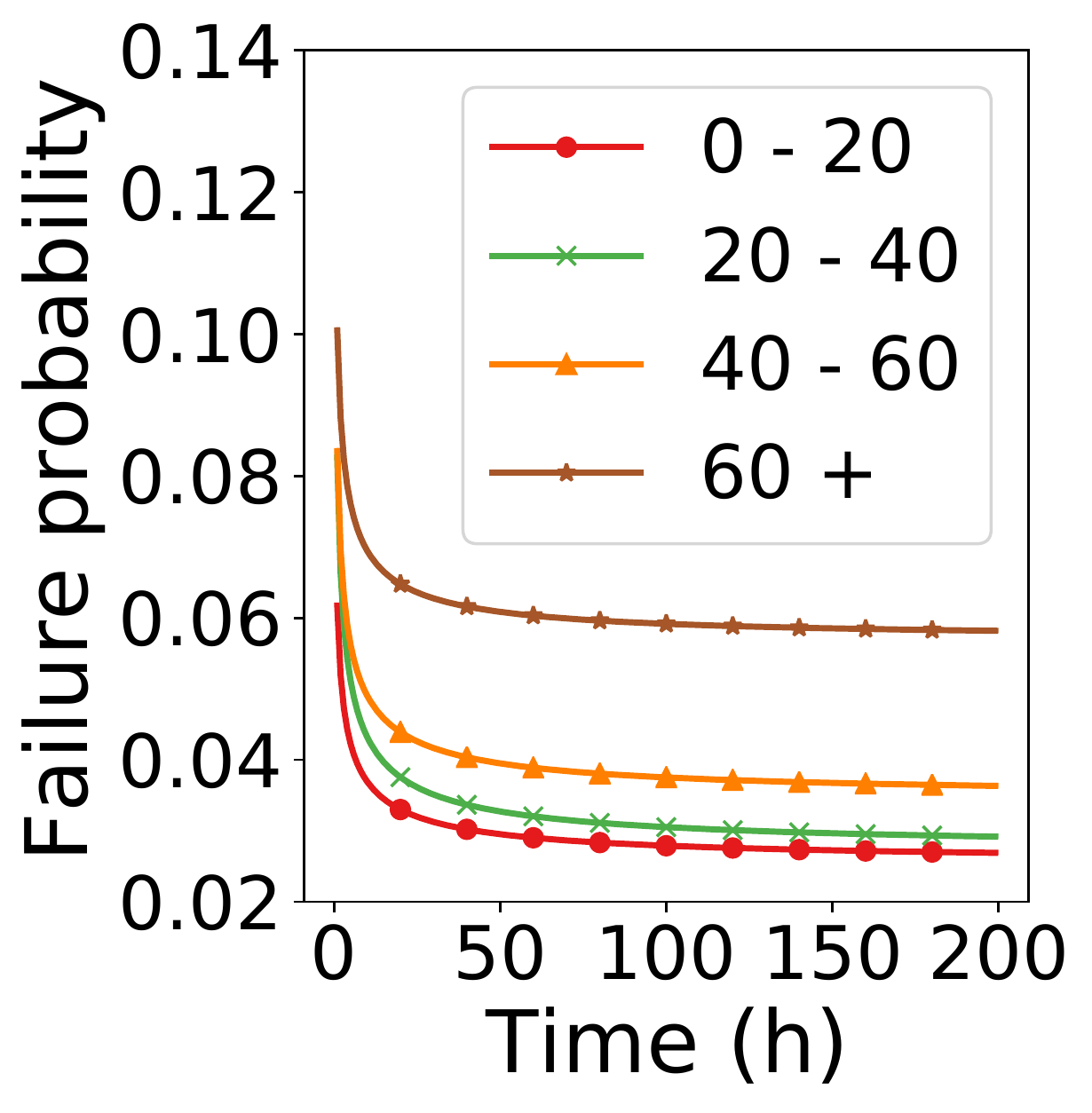}
         \caption{Failure probability.}
         \label{fig:hazard}
     \end{subfigure}
        \vspace{-0.25cm}
        \caption{The observed failure pattern of the training jobs can be fitted as a gamma distribution (a). Corresponding
        failure probability shows a near-constant
        failure probability except at the beginning (b). The label shows the number of participating nodes.}
        \vspace{-0.25cm}
        \label{fig:surv_hazard}
\end{figure}

Failures are common in distributed systems. Prior studies show that the mean-time-between-failures (MTBF) for distributed systems are usually within the order of several hours: 2--27 hours for the high-performance computing cluster of the Los Alamos National Lab~\cite{failure_cmu}; 0.06--58.72 hours from the Google cloud trace log~\cite{failure_google}; 0.08--16 hours for a large-scale heterogeneous server cluster in IBM Research Center~\cite{failure_watson}.

We observed a similar trend in the MTBF with a large collection of recommendation training workflows from the production-scale clusters.
From the logs of 20,000 training jobs running on fleets of Intel 20-core 2GHz processors connected with 25Gbit Ethernet, similar to~\citet{shadowsync} and \citet{intel}, we collected the time-to-failure data. We excluded training runs without failures in the statistics.
Figure~\ref{fig:surv} plots the survival probability of a training job over time.
We overlaid a fitted gamma distribution on top of the observed survival probability data and extrapolated the failure probability, shown in Figure~\ref{fig:hazard}.

The median-time-between-failure (corresponding to $y=0.5$ in Figure~\ref{fig:surv}) was 8--17 hours and the MTBF was 14--30 hours, similar to statistics from prior work~\cite{failure_cmu, failure_fta, failure_google, failure_watson}.
Jobs with more nodes failed more quickly, with the MTBF decreasing {\it linearly} with the increasing number of nodes.
%
Similar to prior work on modeling hardware failures~\cite{four_years}, the observed training failures followed a gamma distribution closely, with an RMSE of 4.4\%. The gamma distribution fits the best compared to other commonly-used heavy-tailed distributions, e.g., Weibull~\cite{weibull}, exponential~\cite{exponential}, and log-normal~\cite{lognormal}.
The derived failure probability was close to uniform, except near the beginning of training (Figure~\ref{fig:hazard}). The much higher failure probability near the beginning is likely related to user errors, e.g., erroneous configuration leading to instant failure.

\subsection{Checkpoint Overhead Analysis for Reliable Recommendation Training}
\label{sec:data_chkpt}

\begin{figure}
     \centering
     \includegraphics[width=0.48\textwidth]{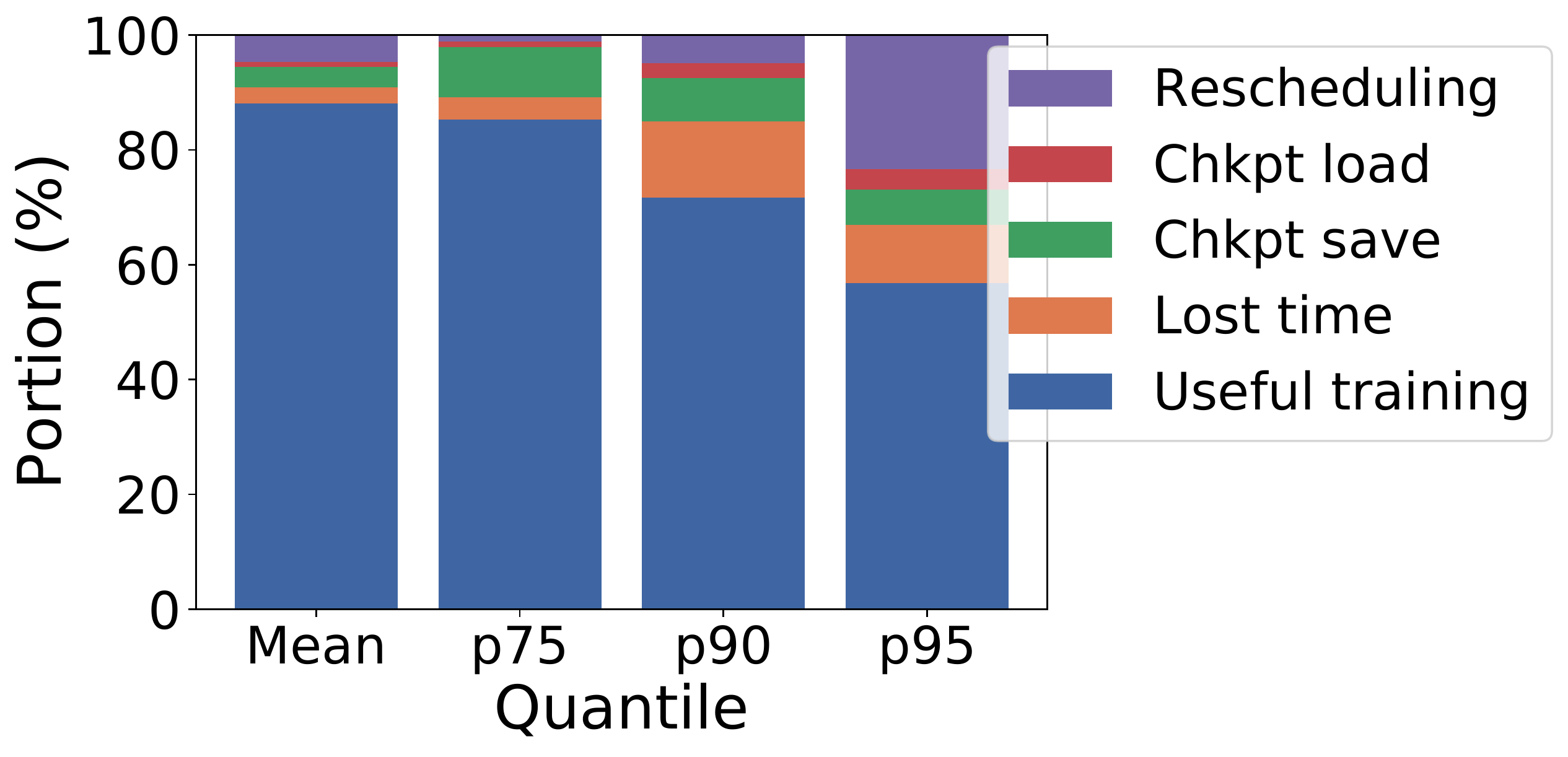}
     \vspace{-0.25cm}
     \caption{Checkpoint-related overheads are responsible for non-negligible amount of training time.
    }
     \vspace{-0.25cm}
     \label{fig:overhead0}
\end{figure}

We quantified the impact of the four checkpoint-related overheads from Section~\ref{sec:bg_chkpt} in a production-scale training cluster.
Similar to the failure analysis, we inspected 17,000 training jobs that ran for more than 10 hours
over a 30-day period.
Figure~\ref{fig:overhead0} shows the
checkpoint-related overhead breakdown.
The four overhead categories added up to an average of 12\% of the total training time.
We estimated the wasted machine-time due to training failures by multiplying the time wasted with the number of nodes.
Even though the average overhead of
12\% may seem small, the implication is dire:
the total overhead of the 17,000 training jobs
summed up to \emph{1,156 machine-year} worth
of computation.

Figure~\ref{fig:overhead0} shows that the overhead is not dominated by a single source.
The major source of overhead for training jobs experiencing fewer failures comes from checkpoint saving (8.8\% for p75), while training jobs with more frequent failures suffered from lost computation (13.2\% for p90) and
rescheduling (23.3\% for p95).
High rescheduling overhead near the tail happens
when the cluster is heavily utilized with additional queuing delay.
The diverse sources of overhead pose a dilemma to full recovery. To optimize for checkpoint saving overheads, a full recovery system must save checkpoints less frequently. However, to optimize for lost computation, the system must save checkpoints more frequently.
Motivated by the dilemma, we explore an alternative solution, partial recovery.
Next section describes the proposed system, \sys, that applies partial recovery to recommendation model training.

\section{The \sys System}
\label{sec:sys}

\begin{figure}
    \centering
    \includegraphics[width=0.46\textwidth]{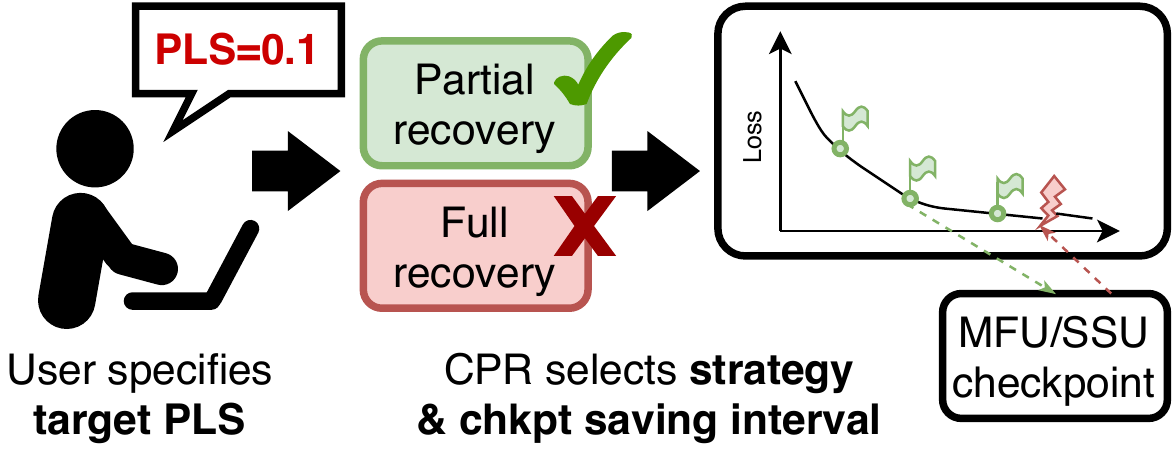}
    \vspace{-0.25cm}
    \caption{\sys selects between full and partial recovery based on the benefit analysis. \sys selects the checkpoint saving interval based on the target \pls. \sys uses MFU/SSU optimization.}
    \vspace{-0.25cm}
    \label{fig:sys}
\end{figure}

\sys is a lightweight design to improve the efficiency and reliability of distributed recommendation model training with partial recovery.
In order to understand the performance--accuracy trade-off space of partial recovery, we propose a new metric---\emph{portion of lost samples} (\pls)---which is a function of checkpoint saving interval, the failure rate of the system, and the number of Emb PS nodes.
Our analysis shows that \pls strongly correlates to the final model accuracy. Using this fact, a \sys user selects a target \pls corresponding to the degree of accuracy degradation that is tolerable. 
\sys selects a checkpoint saving interval to achieve the target \pls. When the selected interval brings too much overhead, \sys simply falls back to full recovery.
To improve the accuracy for \sys further, we add two optimizations, \sys-MFU and \sys-SSU, that prioritize saving embedding vectors with large changes.
Figure~\ref{fig:sys} \sys.

\subsection{Portion of Lost Samples (\pls)}

\pls represents the portion of the training data samples whose effect on the model was lost due to a failure. We empirically show that \pls has a high correlation to final model accuracy and can be used to trade-off performance and accuracy.
Let $S_{total}$ denote the number of total samples, $S_i$ the number of samples processed up to $i$-th iteration, and $N_{emb}$ the number of Emb PS. The \pls at iteration $i$ is:
\begin{equation}
    PLS_{i} = 
    \begin{cases}
      0, & \text{if}\ i=0 \\
      PLS_{i-1} + \frac{S_i - S_{last\_chkpt}}{S_{total} N_{emb}}, & \text{if}\ \text{failure at}\ i \\
      PLS_{i-1}, & \text{otherwise}.
    \end{cases}
    \label{eq:pls}
\end{equation}

$\frac{S_i - S_{last\_chkpt}}{S_{total}}$ represents the portion of the lost samples among total samples.
$N_{emb}$ in the denominator accounts for the fact that the lost information from a node failure is roughly $1/N_{emb}$ with $N_{emb}$ nodes.
As verified below in Section~\ref{sec:eval_pls} with measurement data and analysis, the final model quality is linearly correlated with the final \pls value of the recommendation training system.
Using this relationship, a \sys user can provide a target \pls corresponding to the accuracy degradation they are willing to tolerate. \sys selects the checkpoint saving interval so that the \emph{expected \pls} of the system meets the target \pls.
The \emph{expected \pls} can be calculated from the checkpoint saving interval and the failure frequency:
\begin{equation}
    \mathbb{E}[\pls] = \frac{0.5 T_{save}}{T_{fail}N_{emb}}
\label{eq:pls_exp}
\end{equation}
We briefly describe the derivation.
Expected \pls is the number of expected node failures times the expected \pls increase on each node failures.
With the time at $i$-th iteration of $t_i$,
the total training time of $T_{total}$,
and the expected value for all $i$, $\mathbb{E}_i$,
the expected \pls increases on node failures is:
$\mathbb{E}[\Delta \pls] = \mathbb{E}_{i}[\frac{S_{i} - S_{last\_chkpt}}{S_{total}N_{emb}}]
    = \mathbb{E}_{i}[\frac{t_{i} - t_{last\_chkpt}}{T_{total}N_{emb}}]
    = \frac{0.5T_{save}}{T_{total}N_{emb}}$.
This derivation assumes a constant sample processing rate.
Multiplying this term with the expected number of failures ($\frac{T_{total}}{T_{fail}}$) leads to Equation~\ref{eq:pls_exp}.

With the target \pls specified, \sys can use Equation~\ref{eq:pls_exp} to directly calculate the checkpoint saving interval to use: $T_{save,part} = 2 (\pls) N_{emb} T_{fail}$.
Later, we show in Section~\ref{sec:eval_emul} that the optimal checkpoint saving interval for partial recovery is often much larger than that for full recovery.
The less frequent checkpoint saving of partial recovery brings an additional performance benefit over full recovery.
When the checkpoint saving interval is too small to reap performance benefit from, \sys falls back to full recovery.

\subsection{\sys Execution and Optimization}
\label{sec:sys_cpr}

\paragraph{\pls-based checkpointing.}
\sys selects a checkpoint saving interval using the user-specified target \pls.
The exact relationship between \pls and the final model accuracy depends on the machine learning algorithm and model in use. Selecting a correct target \pls requires empirical knowledge; however, we show in Section~\ref{sec:eval_emul} that choosing a rough, conservative value (e.g., 0.1) works reasonably well across many setups.
After calculating the checkpoint saving interval, \sys compares the estimated overhead of using full recovery (Equation~\ref{eq:overhead}) and partial recovery (Equation~\ref{eq:overhead_par}) using the selected interval to see if partial recovery can bring benefit.
If the expected benefit is small or if there is no benefit, \sys uses full recovery.
Our evaluation shows that our production-scale training cluster can have a large benefit by adopting partial recovery (Section~\ref{sec:eval_emul}).

\begin{figure}
    \centering
    \includegraphics[width=0.48\textwidth]{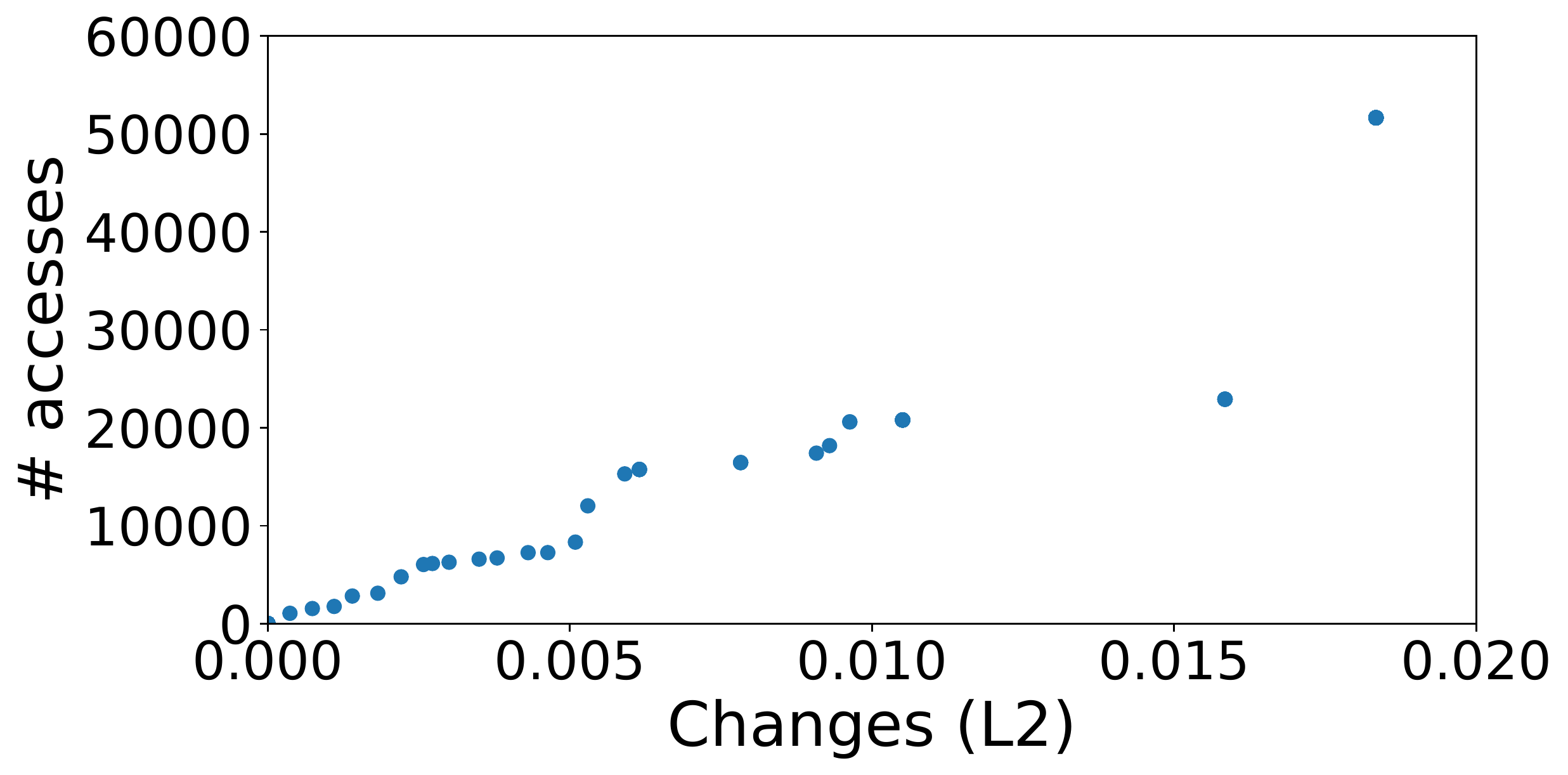}
    \vspace{-0.25cm}
    \caption{The size of the changes has a
    strong correlation (0.9832) with the access
    frequency to the particular embedding
    vectors.}
    \vspace{-0.25cm}
    \label{fig:l2_vs_access}
\end{figure}

\paragraph{Frequency-based prioritization.}
Partial recovery unavoidably loses updates made to the embedding vectors. With the limited I/O bandwidth, prioritizing to save important updates can make the final model quality to improve.
A recent work (SCAR)~\cite{scar} proposed a heuristic to prioritize saving parameters with larger changes.
By tracking the $L^2$-norm of the updates, SCAR saves the most-changed $rN$ parameters every $rT_{save}$ $(r<1)$, instead of saving $N$ parameters every $T_{save}$.

\sys can potentially benefit from adopting SCAR.
However, SCAR is impractical to implement in industry-scale recommendation training systems. 
Tracking updates to the embedding tables of several TBs in size requires the same order-of-magnitude memory capacity, at most requiring as much memory as the model itself. 
Furthermore, selecting the top $rN$ most changed 
vectors has a time complexity of $O(N \log(N))$, scaling poorly with increasing $N$.

Instead of tracking the updates directly, we propose to only track the access frequency.
Figure~\ref{fig:l2_vs_access} shows the strong
correlation between the access frequency
and the size of the update to embedding vectors, measured after 4096 iterations for the Kaggle dataset~\cite{kaggle} (evaluation details in Section~\ref{sec:eval}).
The correlation coefficient is high, at 0.983, meaning that the access frequency is an excellent proxy to the magnitude of the embedding vector update.
Based on this observation, we propose time- and memory-efficient alternatives over SCAR: 
\sys-MFU and \sys-SSU.

\paragraph{\sys-MFU.}
\sys-MFU saves the Most-Frequently-Used (MFU) $rN$
out of $N$ parameters on every $rT_{save}$,
with $r < 1$.
A 4-byte counter is allocated for each vector in the embedding table to track the access frequency. The typical size of an embedding vector ranges from 64--512 bytes~\cite{dlrm}, making the memory
overhead of the counter 0.78--6.25\% of the size of the embedding tables.
This is much smaller compared to the 100\% memory overhead of SCAR.
When an embedding vector is saved, its counter is cleared.
The time complexity, however, is the same with SCAR, being in the order of $O(N \log(N))$.

%
%

\paragraph{\sys-SSU.}
\sys-SSU further improves the time and memory overhead of \sys-MFU.
\sys-SSU \emph{Sub-Samples} Used (SSU) embedding vectors and keeps a list of vectors that were ever accessed from the subsampled data points, of size $rN$.
If the list overflows, \sys-SSU randomly discards the overflowing entries.
The idea of \sys-SSU is that the subsampling will
act as a high-pass filter, giving vectors with more frequent accesses a higher likelihood of staying in the list.
Because \sys-SSU only requires
a list of size $rN$, the memory overhead is
$r < 1$ times that of \sys-MFU.
With $r=0.125$, the memory overhead becomes 0.097--0.78\% of the embedding tables.
\sys-SSU only needs to keep a non-duplicate list of size
$rN$, which has a time complexity in the order
of $O(N)$.
Table~\ref{tbl:mfu_ssu} summarizes the overhead
of SCAR, \sys-MFU, and \sys-SSU.

\begin{table}[t]
    \centering
    \begin{tabular}{c|c|c}
         & Time & Mem (rel. to emb tbl)\\\hline
    SCAR & $\approx O(Nlog(N))$ & $100\%$\\
    MFU & $\approx O(Nlog(N))$ & $0.78-6.25\%$\\
    SSU & $\approx O(N)$ & $0.097-0.78\%$
    \end{tabular}
    \caption{Time and memory overhead of SCAR~\cite{scar}, \sys-MFU, and \sys-SSU.
    Memory overhead is shown for embedding vectors of size 64--512 bytes, with $r=0.125$.}
    \label{tbl:mfu_ssu}
\end{table}
\section{Experimental Methodology}

We evaluated \sys in two different settings: (1) a framework that emulates the characteristics of the production-scale cluster, and (2) a real production-scale cluster.

\subsection{Emulation Framework}

The emulation framework allows a fast evaluation of \sys using a small model and a small dataset, while emulating the failure/overhead characteristics from the production cluster.
For emulation, we trained the DLRM recommendation architecture~\cite{dlrm}, a standard reference model for implementing recommendation systems in MLPerf~\cite{mlperf}. We directly implemented \sys on top of the reference implementation provided by MLPerf.
We trained DLRM using two datasets of different sizes, the Criteo Kaggle~\cite{kaggle} and Terabyte datasets~\cite{terabyte}, which are both reference datasets used in MLPerf~\cite{mlperf}.
The hyperparameters of DLRM differed depending on the dataset, following the reference implementation from MLPerf~\cite{mlperf}.
For the Kaggle dataset, we use 64-byte embedding vectors, a 4-layer Bottom MLP of (13$\times$512, 512$\times$256, 256$\times$64, 64$\times$16), and a 3-layer Top MLP of (432$\times$512, 512$\times$256, 256$\times$1).
For the Terabyte dataset, we use 256-byte embedding vectors, a 3-layer Bottom MLP of (13$\times$512, 512$\times$256, 256$\times$64),
and a 4-layer Top MLP of (1728$\times$512, 512$\times$512, 512$\times$256, 256$\times$1).
DLRM was trained on a single machine with two NVIDIA V100 GPUs attached to a server with 20 CPUs and 64GB memory.
Using a single node does not affect the accuracy of DLRM because the implementation is fully synchronous.
Our implementation of \sys on top of DLRM did not increase the training time meaningfully.

We ran training for a single epoch using all the data samples and reported the final test \emph{receiver operating characteristic area under curve} (AUC). AUC is less susceptible to unbalanced datasets and is a standard metric for evaluating DLRM~\cite{mlperf}.
Training for a single epoch is common for DLRM~\cite{dlrm,mlperf-mlsys,mlperf-ieeemicro}, because DLRM suffers from overfitting if the same training data is revisited.

\paragraph{Failure and overhead emulation.}
Because the emulation runs much faster than production-scale training, we \emph{project} the failure/overhead characteristics from Section~\ref{sec:analysis} down to account for the training time difference.
We emulate a \emph{56-hour} training job for simplicity; the average number of failures for a 56-hour training was exactly 2.
We inject 2 failures randomly, as the failure probability is nearly uniform for the real-world cluster (Section~\ref{sec:bg_failure}).
A failure clears 50\%, 25\%, or 12.5\% of the embedding tables and triggers partial recovery, emulating a portion of the Emb PS failing.
%
%
%
We linearly scale down the checkpoint-related overheads and the checkpoint saving interval.
%

\paragraph{Strategies.}
We implemented and compared full recovery, baseline partial recovery, \sys-vanilla, \sys-SCAR, \sys-MFU, and \sys-SSU.
Full recovery uses the optimal checkpoint saving interval ($T_{save} = \sqrt{2 O_{save}T_{fail} }$). The baseline partial recovery naively uses the same interval. 
\sys calculates the checkpoint saving interval from the target \pls.
We compare four different variants of \sys: \sys-vanilla calculates the checkpoint saving interval from the target \pls without additional optimizations. \sys-SCAR implements the SCAR optimization from prior work~\cite{scar}, which imposes significant memory overhead.
\sys-MFU/SSU applies our memory-efficient MFU/SSU optimizations.
For \sys-SSU, we use a sampling period of 2.
We only apply SCAR/MFU/SSU optimizations to the
7 largest tables among 26 tables, which
take up 99.6\% (Kaggle) and 99.1\% (Terabyte) of the entire table size, respectively.
For the 7 tables, we save checkpoints
8 times frequently but at each only save at most
1/8 of the parameters compared to full recovery (i.e., $r=0.125$).
Other tables are always fully saved.

\begin{figure}
    \centering
    \includegraphics[width=0.46\textwidth]{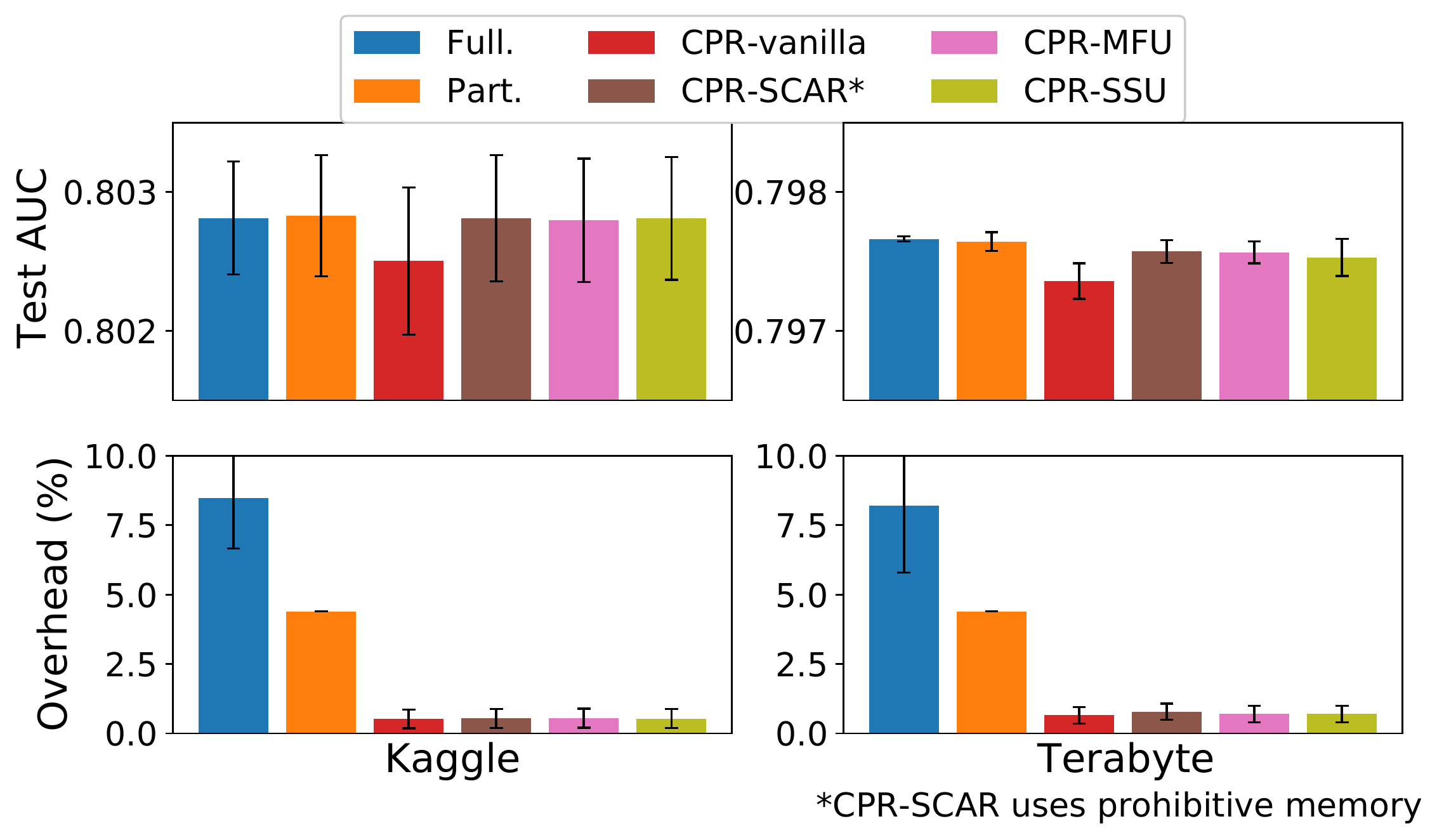}
    \caption{\sys reduces the checkpoint-related overhead over full recovery by 93.7\% (Kaggle) and 91.7\% (Terabyte) on a setup emulating the production cluster, while achieving similar accuracy.
    }
    \label{fig:perf_acc}
\end{figure}

\subsection{Production-scale Cluster}

The evaluation on the production-scale cluster used 20 MLP trainers and 18 Emb PS nodes.
Each node consists of Intel 20-core, 2GHz processors connected with 25Gbit Ethernet, similar to~\cite{shadowsync}.
We trained the model for 10 hours, during which one failure was injected.
To simply test the effect of partial recovery, we mimicked the behavior of partial recovery by switching part of the checkpoints to an older version and triggering full recovery right after saving a checkpoint.
Except for the nodes whose checkpoints were switched, loading checkpoints would not revert the model, effectively having the same effect as partial recovery.


\section{Evaluation Results}
\label{sec:eval}

\subsection{Emulation Results: Training Time and Accuracy}
\label{sec:eval_emul}

We ran full recovery (Full.), baseline partial recovery (Part.), and different variants of \sys on our failure emulation framework, which closely emulates our production-scale cluster's failure rate and the checkpoint saving overhead. For the variants of \sys, we used target \pls=0.1.
Figure~\ref{fig:perf_acc} summarizes the result for both Kaggle and Terabyte datasets.

\paragraph{\sys reduces the training time.}
Compared to full recovery, \textbf{\sys reduces the checkpoint-related overhead by 93.7\% and 91.7\%} for Kaggle and Terabyte dataset, respectively.
The speedup can be broken down into two factors. Elimination of the lost computation reduces the overhead both from 8.5\% (Kaggle) and 8.2\% (Terabyte) to 4.4\% (Full. vs. Part.). \pls-based checkpoint saving interval selection additionally brings down the 4.4\% to 0.53\% and 0.68\% for Kaggle and Terabyte dataset, respectively (Part. vs. CPR).

\begin{figure}
    \centering
    \includegraphics[width=0.49\textwidth]{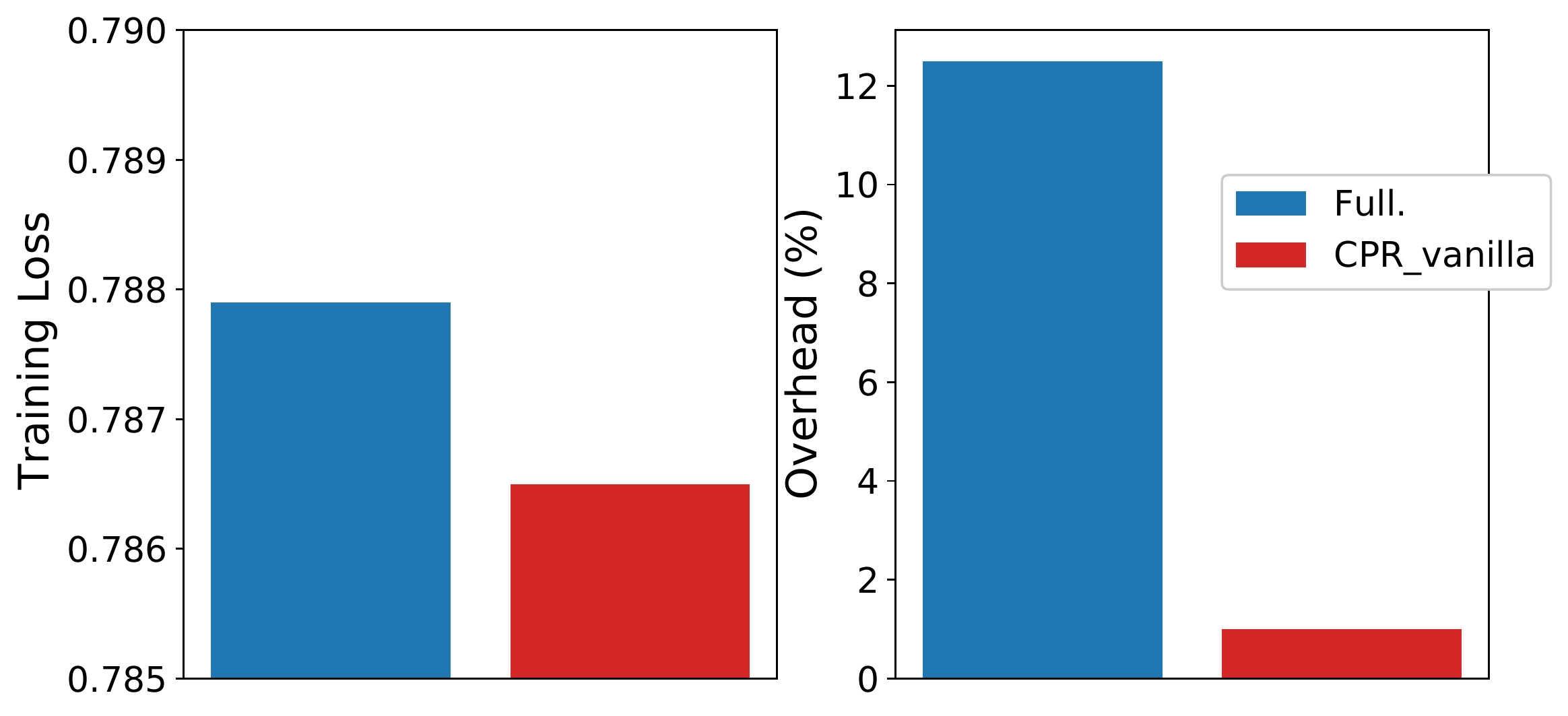}
    \caption{\sys practically showed the same accuracy while reducing the 12.5\% overhead to 1\% on a production-scale setup.}
    \label{fig:perf_acc_prod}
\end{figure}

\paragraph{\sys maintains reasonable accuracy.}
With optimizations, \textbf{\sys was able to mostly achieve accuracy on par with full recovery, losing at most only 0.0002 test AUC} with optimizations.
For both full recovery and baseline partial recovery, the test AUC was 0.8028/0.7977 for Kaggle/Terabyte dataset.
\sys-vanilla trades off accuracy with performance.
While reducing the overhead to a marginal 0.53--0.68\%, the AUC for \sys-vanilla decreased to 0.8025/0.7974, for Kaggle/Terabyte dataset (0.04\% degradation).
%
%
%
\sys-SCAR/MFU/SSU improves accuracy, making \sys much more practical. For Kaggle, they all reached a test AUC on par with that of full recovery.
For Terabyte, \sys-SCAR/MFU/SSU achieved AUC=0.7976/0.7976/0.7975 (0.011/0.012/0.017\% degradation), respectively.
While using less memory (Table~\ref{tbl:mfu_ssu}) \sys-MFU/SSU achieved accuracy similar to that of \sys-SCAR.

\subsection{Production-scale cluster Results: Training Time and Accuracy}

We evaluated full recovery and \sys-vanilla on a production-scale cluster. We injected a single failure near the end of the training that failed 25\% of the Emb PS nodes.
\sys-vanilla used a target \pls of 0.05 and a resulting checkpoint saving interval of 4 hours.
Full recovery saved checkpoints every 2 hours.
Because the production-scale training did not report AUC, we report the training loss instead.

Figure~\ref{fig:perf_acc_prod} summarizes the result.
The accuracy did not degrade with \sys-vanilla--it actually improved slightly, although the difference was small.
Meanwhile, the overhead decreased significantly with \sys-vanilla, from 12.5\% to 1\%. Most of the overhead reduction (10\%) came from the elimination of lost computation. 1.5\% reduction came from saving checkpoints less frequently.
The limited number of data points suggests a possible benefit of using \sys in a production environment.
We did not evaluate \sys-MFU/SSU, because the accuracy was already good.

\subsection{Sensitivity Study: \pls}

\begin{figure}
    \centering
    \includegraphics[width=0.48\textwidth]{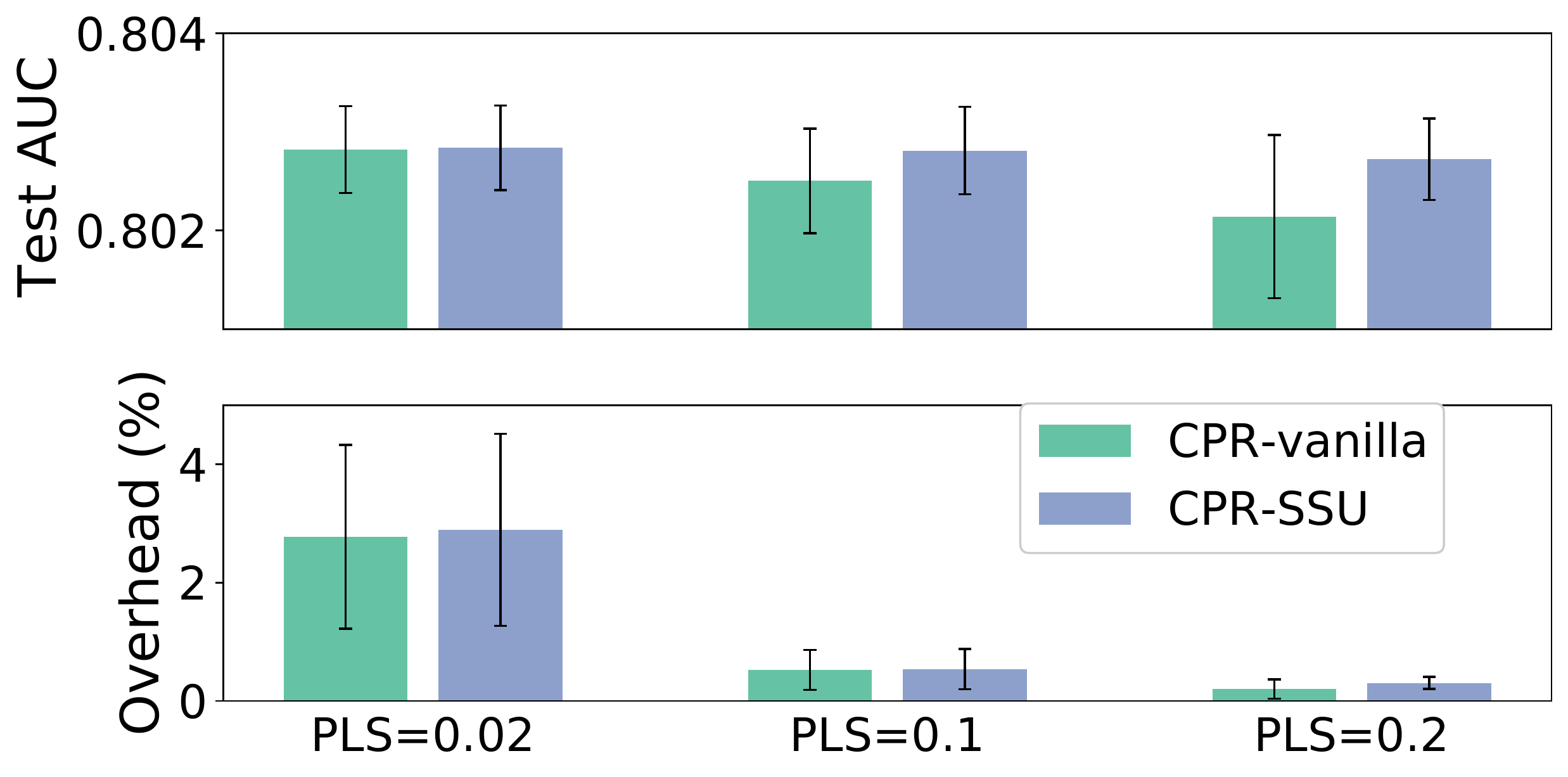}
    \caption{\sys trades off performance and accuracy.}
    \label{fig:sensitivity0}
\end{figure}

To evaluate the effect of different target \pls,
we varied the target \pls between 0.02, 0.1, and 0.2 and present the resulting accuracy and overhead.
We only show the result of \sys-vanilla and \sys-SSU from the Kaggle dataset for brevity; other configurations showed a similar trend.
Figure~\ref{fig:sensitivity0} summarizes the result. For both \sys-vanilla and \sys-SSU, varying \pls effectively traded off accuracy and performance. For \sys-vanilla, increasing the target \pls from 0.02 to 0.2 decreased the overhead from 2.9\% to 0.3\%, while degrading accuracy from AUC=0.8028 to AUC=0.8021.
For \sys-SSU, the degradation was much lower. \sys-SSU experiences a marginal AUC decrease from AUC=0.8028 to AUC=0.8027 for the same speedup.

\subsection{Sensitivity Study: Failures}

We also varied the number of failures and the portion of lost nodes on each failure.
We fixed the target \pls to 0.02.
We varied the number of failures between 2, 20, and 40. 20 and 40 failures represent a hypothetical case where the system experiences 10--20$\times$ more failures.
Such a setup can represent a scenario of off-peak training, a training that only uses idle resources and gets suspended whenever a higher priority job arrives (e.g., Amazon Spot~\cite{spot}).
On each failure, we varied the portion of the Emb PS nodes failed between 12.5--50\%.
We only plot the overhead; the accuracy was similar across all experiments.
The overhead is normalized to the overhead of full recovery for simple comparison.
Again, we only selectively show full recovery and \sys-SSU, trained with Kaggle dataset. Omitted data points showed a similar trend.
The configurations \sys found as not beneficial to run a partial recovery are marked in a red hatch. We still plot what the overhead would have been like had \sys run partial recovery in such setups.

\begin{figure}
    \centering
    \includegraphics[width=0.48\textwidth]{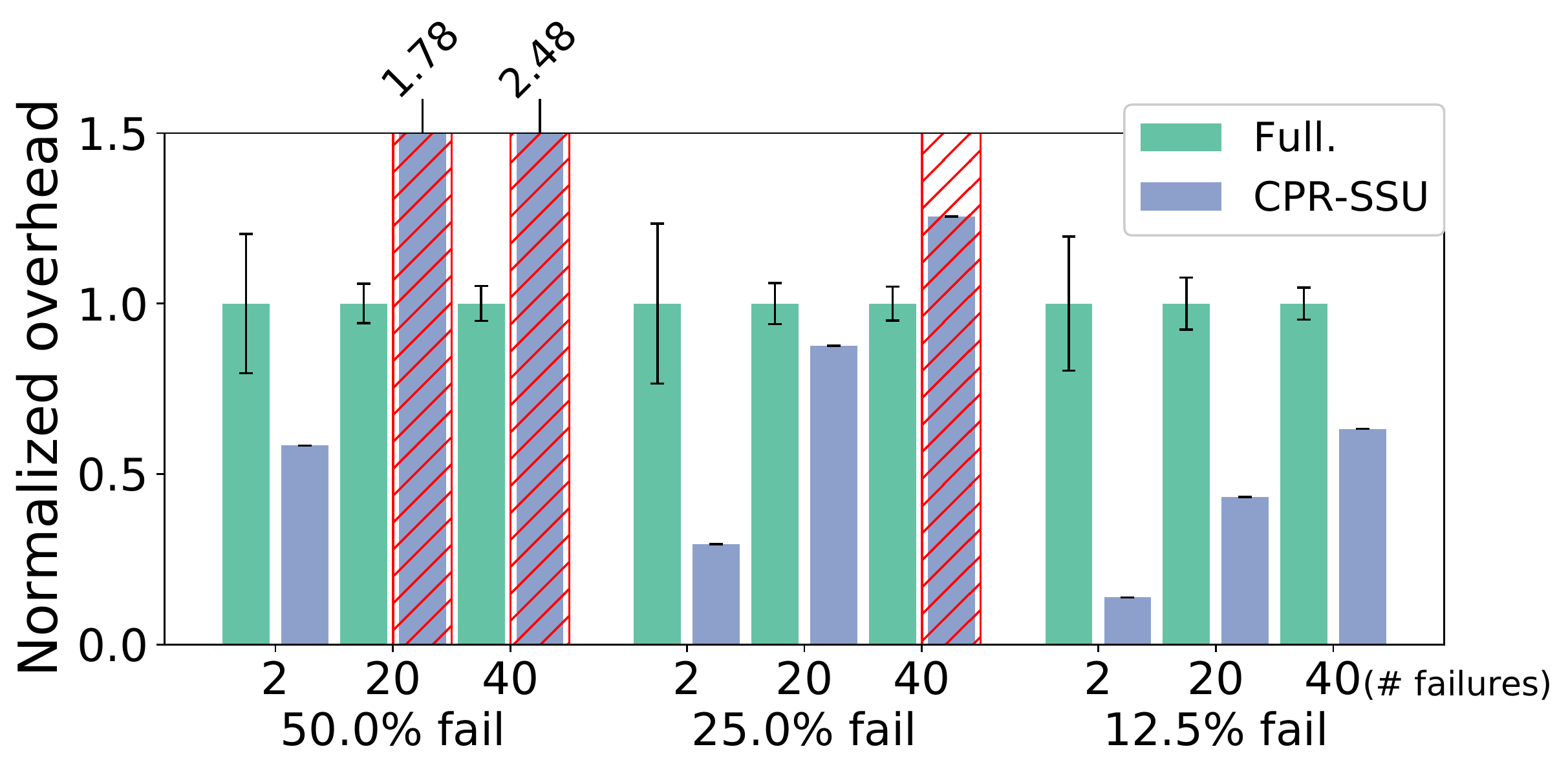}
    \caption{\sys is less effective with more failures.}
    \label{fig:sensitivity1}
\end{figure}

Figure~\ref{fig:sensitivity1} shows that \sys correctly estimates the benefit of using partial recovery.
The overhead of the setup \sys predicted as not beneficial to run partial recovery (red hatch) was all higher than that of full recovery.
Figure~\ref{fig:sensitivity1} also shows that
\sys's speedup becomes smaller when failures occur more frequently or when more nodes fail at once.
%
%
%
\sys is less effective with more frequent failures because the checkpoint saving interval of partial recovery ($2 (PLS) N_{emb} T_{fail}$) decreases faster with decreasing mean-time-to-failure, compared to full recovery ($\sqrt{2 O_{save} T_{fail}}$).

\subsection{PLS and Accuracy}
\label{sec:eval_pls}

\begin{figure}
     \centering
     \begin{subfigure}[b]{0.225\textwidth}
         \centering
         \includegraphics[width=\textwidth]{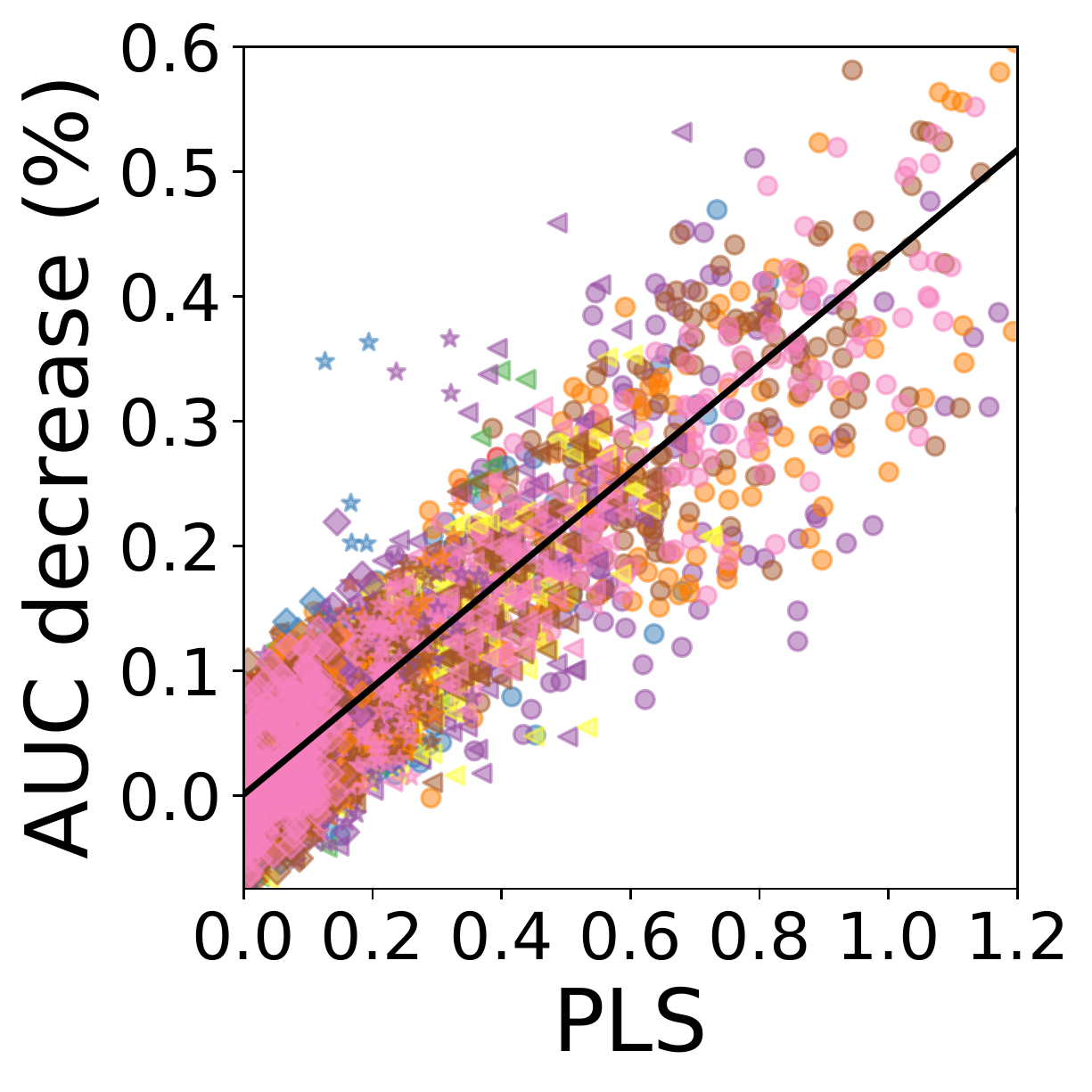}
         \caption{Kaggle dataset.}
         \label{fig:scatter_kaggle}
     \end{subfigure}
     \begin{subfigure}[b]{0.23\textwidth}
         \centering
         \includegraphics[width=\textwidth]{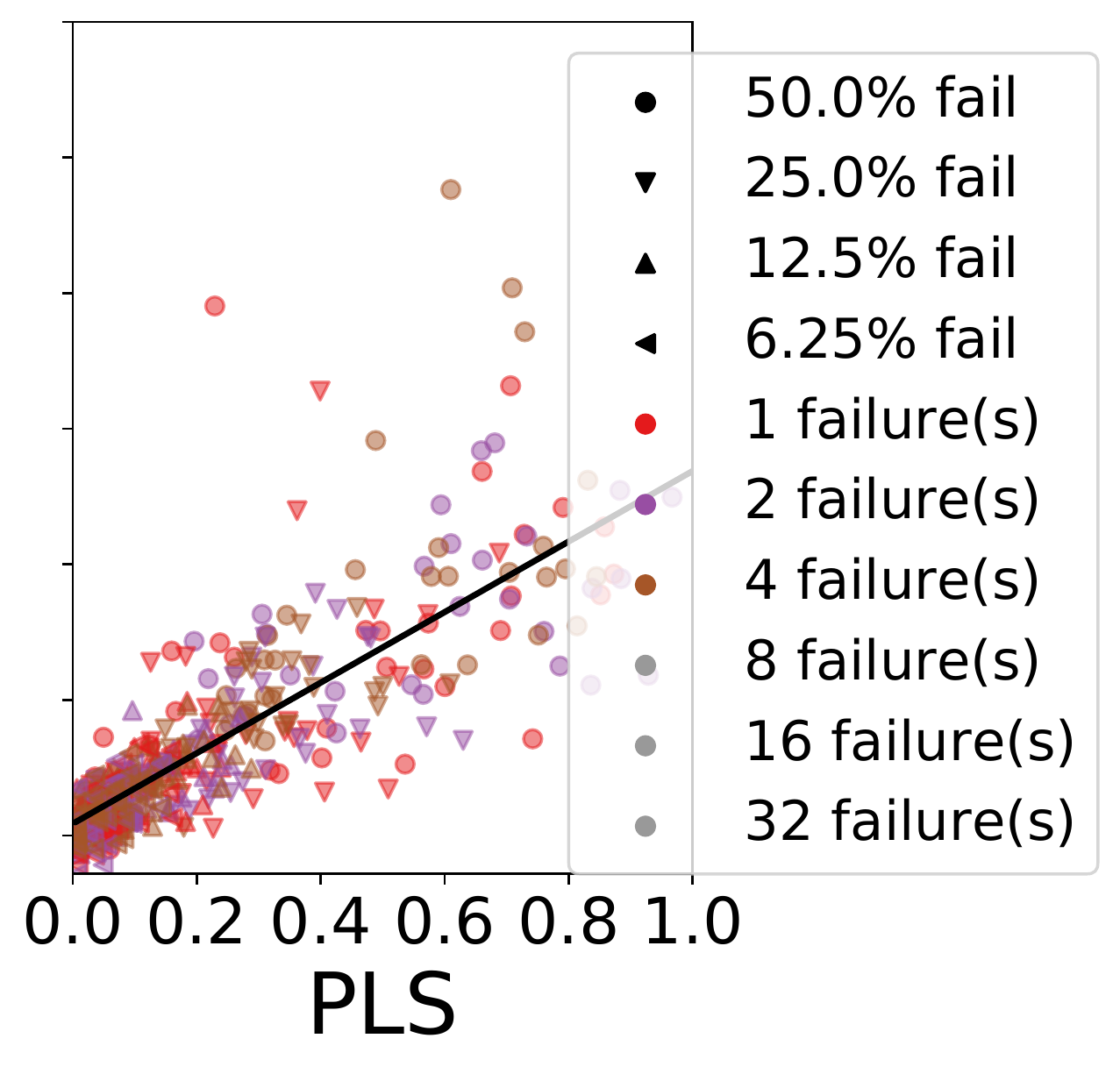}
         \caption{Terabyte dataset.}
         \label{fig:scatter_tb}
     \end{subfigure}
        \caption{
        \pls shows a strong
        correlation with the model
        accuracy for both
        Kaggle (corr=0.8764) and Terabyte (0.8175) dataset.
        }
        \label{fig:scatter_uc}
\end{figure}

\begin{figure}
    \centering
    \includegraphics[width=0.49\textwidth]{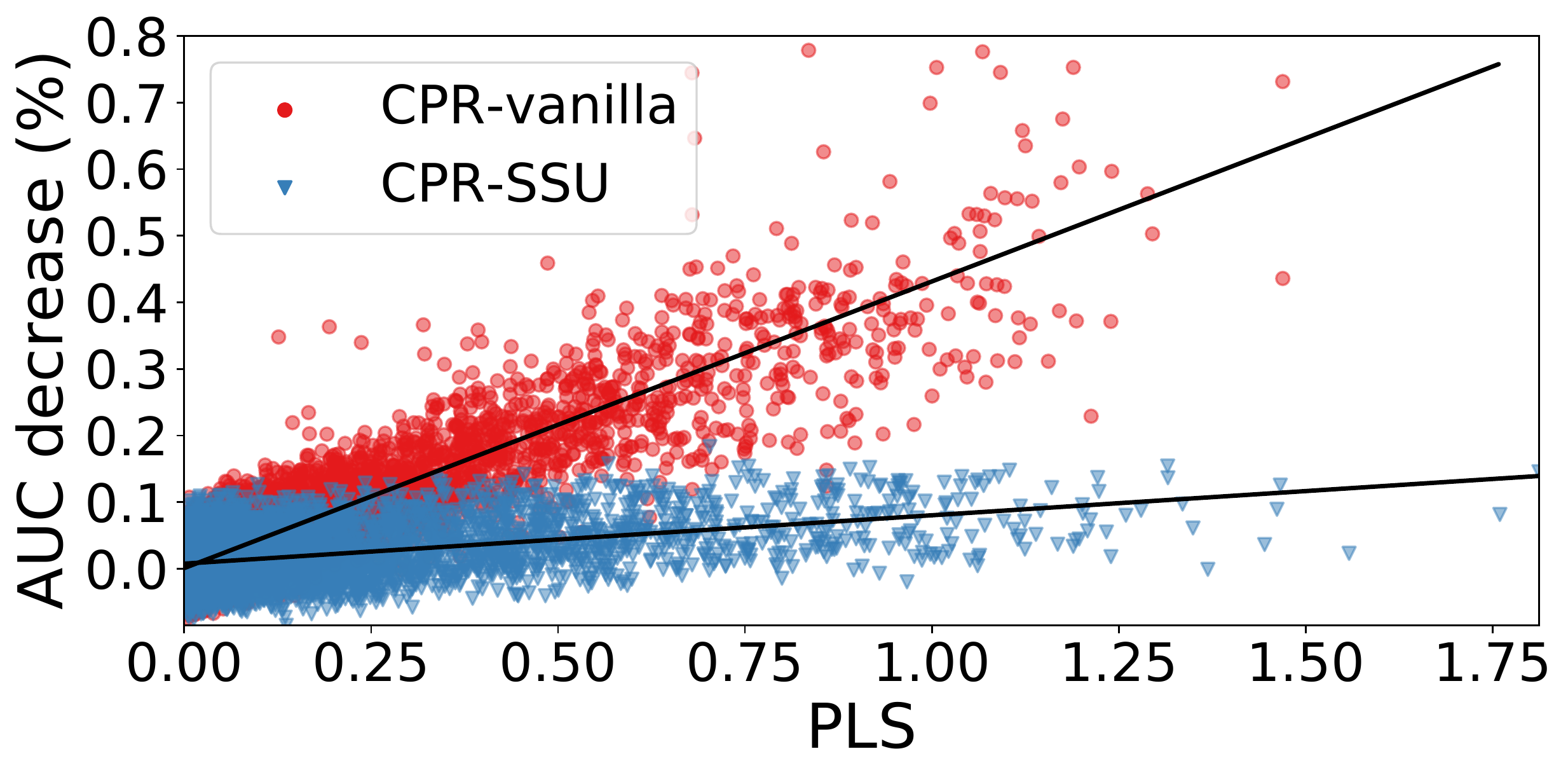}
    \caption{
    \sys-SSU (blue) reduces the slope for the \pls--accuracy relationship compared to \sys-vanilla (red), allowing \sys to expand the useful range of \pls values.}
    \label{fig:scatter_cpr}
\end{figure}

\sys relies on the linear relationship between the \pls and the final model accuracy.
To evaluate the relationship,
we generated runs with 1--32 random failures, each clearing 6.25--50\% of the embedding vectors. We also randomly selected a checkpoint saving interval so that the expected \pls falls near 0--1. We applied partial recovery without any optimization and plotted the final accuracy degradation compared to the non-failing case.

Figure~\ref{fig:scatter_uc} shows the strong linear relationship between \pls and the final model accuracy.
The correlation holds regardless of the number of the failed portion or the failure frequency, i.e., there is no strong correlation between the failure frequency or the failed portion and the final accuracy as long as the \pls values are the same.
With the relationship known in prior, a \sys designer can limit the accuracy degradation by specifying a target \pls.
Figure~\ref{fig:scatter_cpr} additionally plots the \pls--accuracy relationship for \sys-SSU.
The figure shows that
\sys-SSU reduces the slope significantly, enabling \sys to explore a larger range of target \pls.

%

\subsection{Partial Recovery Scalability Analysis}
\label{sec:eval_scalability}

\begin{figure}
     \centering
     \begin{subfigure}[b]{0.23\textwidth}
         \centering
         \includegraphics[width=\textwidth]{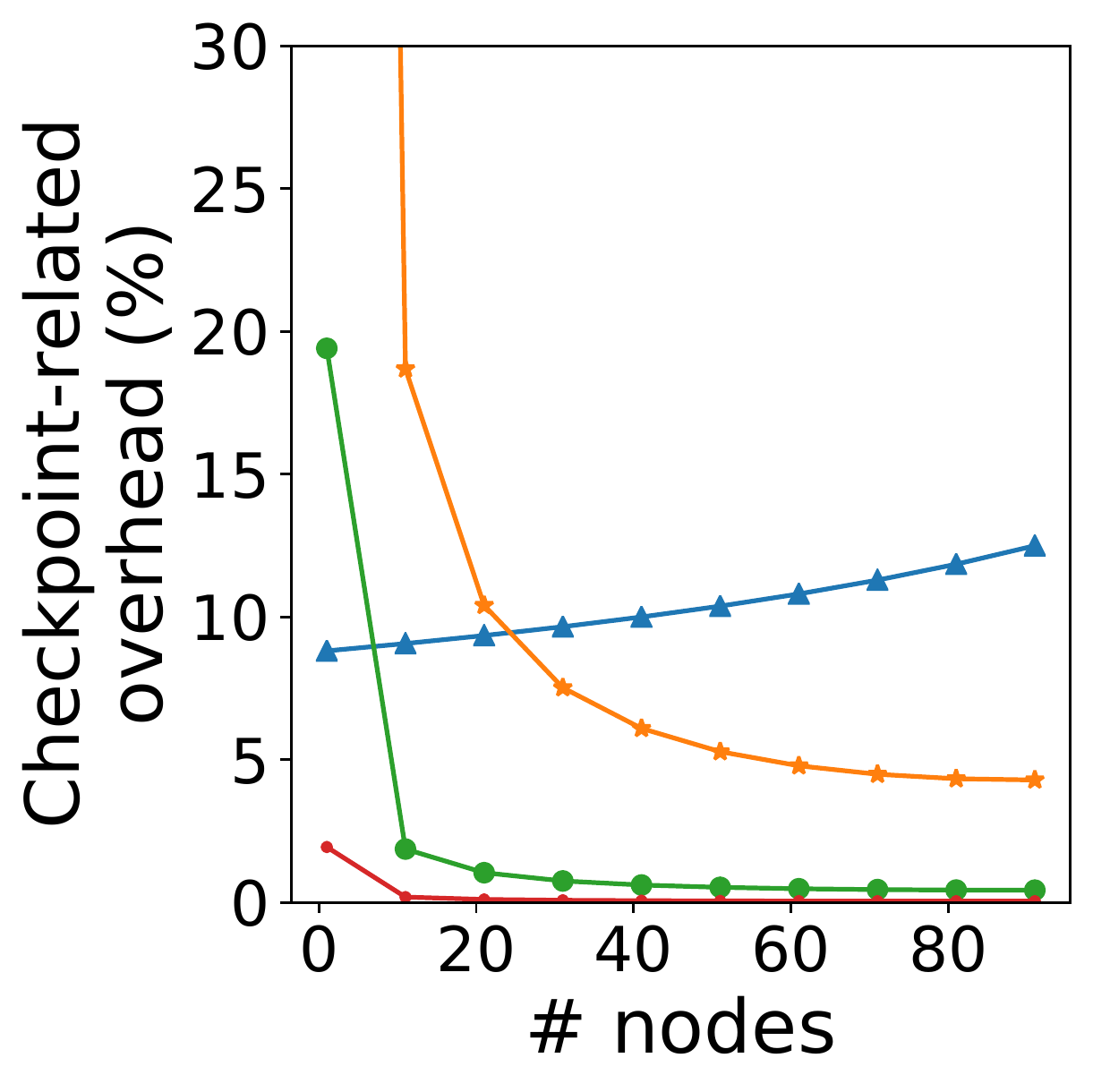}
         \caption{Linear MTBF.}
         \label{fig:scalability0}
     \end{subfigure}
     \begin{subfigure}[b]{0.23\textwidth}
         \centering
         \includegraphics[width=\textwidth]{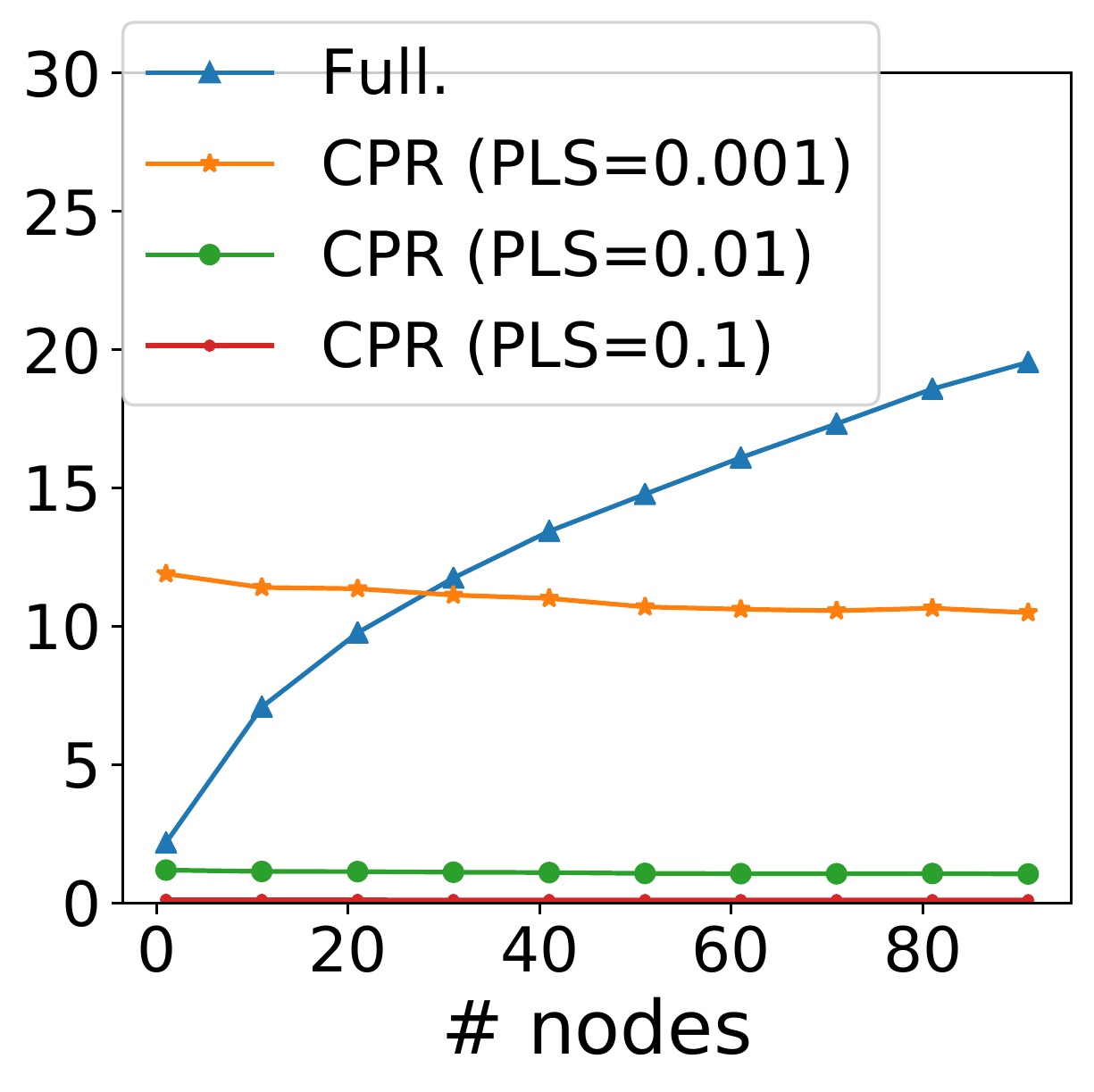}
         \caption{Independent node failures.}
         \label{fig:scalability1}
     \end{subfigure}
        \caption{\sys shows better scalability than full recovery for different failure models.}
        \label{fig:scalability}
\end{figure}

To study the scalability of \sys, we analytically estimated the overhead of full recovery and \sys using Equation~\ref{eq:overhead} and Equation~\ref{eq:overhead_par}.
To conjecture how the rate of node failures would increase, we assumed two different models: (1) linearly decreasing mean-time-between-failure (MTBF) with an increasing number of nodes, which was the behavior observed from Section~\ref{sec:bg_failure}, and (2) assuming that each node has an independent failure probability $p$. The second model leads to an MTBF equation in the form of $\frac{1}{1-(1-p)^{n}}$, which deviates from the linear behavior seen from the production cluster. Still, we consider this model due to its simplicity.

Figure~\ref{fig:scalability} plots the result. For both of the failure models, \sys showed better scalability than full recovery, where the overhead actually \emph{decreased} with an increasing number of nodes. For both cases, full recovery saw an \emph{increasing} overhead with the increasing number of nodes.
\sys scales better with an increasing number of nodes because, although the probability of observing a failure increases, the \emph{portion of the updates lost decreases} with the number of nodes. Full recovery loads all the checkpoints even if only a small fraction of the model is lost, resulting in worse scalability.
\section{Additional Related Work}
\label{sec:related}
\subsection{Prior Work on Checkpointing}

\paragraph{Checkpointing for non-ML applications.}
Checkpointing is a common technique
used in data centers to handle failures~\cite{lamport, checkpointing}.
Traditional checkpointing saves a globally
consistent state across all the participating nodes and use full recovery to ensure correct behavior~\cite{lamport}, which is often expensive.
Many optimizations orthogonal to \sys have been
proposed to speed up checkpointing, including using multi-level hierarchy~\cite{scr, fti}, adding
asynchronous operations~\cite{veloc},
leveraging memory access patterns~\cite{aickpt, carbone}, or
simultaneously using logging~\cite{lineagestash}.
These works aim to support arbitrary workloads and
are complementary.

Intermittent computing~\cite{mementos, quickrecall, chinchilla, alpaca, samoyed, catnap, ufop, flicker, dino, ratchet, clank, nvp, cospec}, a field 
enabling compute on an energy-harvesting
device with frequent failures, has also widely adopted checkpointing.
However, unlike \sys, these works mainly focus on a single-node system.

\paragraph{Checkpointing for distributed ML training.}
Several distributed training systems implement checkpointing~\cite{adam, pipedream, lazytable}.
Orpheus~\cite{orpheus} incrementally
saves a checkpoint by breaking the
model into a running sum of decomposed
vectors, from which the original model
can be recalculated.
DeepFreeze~\cite{deepfreeze} improves
the checkpointing efficiency by introducing multi-level storage, sharding the work across nodes, and overlapping compute with checkpoint saving.
While some of the prior works reduce checkpoint-related overhead by leveraging ML-specific characteristics,
they do not use partial recovery like \sys.
SCAR~\cite{scar} is the first system that explores the benefit of partial recovery.
\sys additionally studies the trade-off of partial recovery that SCAR neglected and proposes memory-efficient optimizations.

\section{Conclusion and Future Work}
\label{sec:conclusion}


Training a recommendation system requires a fleet
of machines due to its high compute and memory demand.
With the ever-increasing number of participating nodes,
the training process experiences more and more frequent
failures.
We studied the failure characteristics and the
resulting checkpoint-related overheads and observed
that traditional full recovery adds unnecessary
checkpoint saving overhead and lost computation.

We propose \sys, a system leveraging partial recovery
to reduce the checkpoint-related overheads for
recommendation system training.
\sys selects checkpoint saving interval based on the user-specified target \pls, maximizing performance while maintaining reasonable accuracy.
\sys also implements low-overhead optimizations
that further reduce the accuracy degradation.
We show that \sys can effectively eliminate
checkpoint-related overhead with partial recovery
while suppressing significant accuracy degradation.

Partial checkpoint recovery after a failure perturbs the training process. Consequently, when training with CPR it may be beneficial to use more robust distributed training methods, such as those designed to handle more adversarial Byzantine failures~\cite{yin2018byzantine,chen2018draco}. We leave this line of investigation to future work.




\section*{Acknowledgements}
The authors would like to thank Hsien-Hsin Sean Lee, Kim Hazelwood, Udit Gupta, David Brooks, Maxim Naumov, Dheevatsa Mudigere, Assaf Eisenman, Kiran Kumar Matam, and Abhishek Dhanotia
for the discussion and feedback on this work.

\bibliography{bib.bib}
\bibliographystyle{mlsys2020}

\appendix


\end{document}